# Fine-Tuning Florence2 for Enhanced Object Detection in Unconstructed Environments: Vision-Language Model Approach

**Aysegul Ucar[1,*], Soumyadeep Ro[2], Sanapala Satwika[3], Pamarthi Yasoda Gayathri[4], and Mohmmad Ghaith Balsha[5]**

[1] Firat University, Mechatronics Engineering Department, Elazig, Türkiye; agulucar@firat.edu.tr
[2] Indian Institute of Technology Kharagpur, West Bengal, India; soumyadeep1311@gmail.com
[3] Indian Institute of Technology Kharagpur, West Bengal, India; satwikasanapala@gmail.com
[4] Indian Institute of Technology Kharagpur, West Bengal, India; yasodagayatri2002@gmail.com
[5] Firat University, Mechatronics Engineering Department, Elazig, Türkiye; 241134102@firat.edu.tr

\* Correspondence: agulucar@firat.edu.tr

**Abstract:** Vision-Language Models (VLMs) have emerged as powerful tools in artificial intelligence, capable of integrating textual and visual data for a unified understanding of complex scenes. While models such as Florence2, built on transformer architectures, have shown promise across general tasks, their performance in object detection within unstructured or cluttered environments remains underexplored. In this study, we fine-tuned the Florence2 model for object detection tasks in non-constructed, complex environments. A comprehensive experimental framework was established involving multiple hardware configurations (NVIDIA T4, L4, and A100 GPUs), optimizers (AdamW, SGD), and varied hyperparameters including learning rates and LoRA (Low-Rank Adaptation) setups. Model training and evaluation were conducted on challenging datasets representative of real-world, disordered settings. The optimized Florence2 models exhibited significant improvements in object detection accuracy, with Mean Average Precision (mAP) metrics approaching or matching those of established models such as YOLOv8, YOLOv9, and YOLOv10. The integration of LoRA and careful fine-tuning of transformer layers contributed notably to these gains. Our findings highlight the adaptability of transformer-based VLMs like Florence2 for domain-specific tasks, particularly in visually complex environments. The study underscores the potential of fine-tuned VLMs to rival traditional convolution-based detectors, offering a flexible and scalable approach for advanced vision applications in real-world, unstructured settings.

## 1. Introduction

Vision-Language Models (VLMs) represent a major step forward in the field of Artificial Intelligence (AI) [1, 2]. These models enhance AI's ability to comprehend and interact with the environment by combining text and visual information. The VLMs are capable of performing a wide range of tasks that require an understanding of both language and images. They are used in various applications, including object detection, multimodal categorization, visual question and answering, and image captioning. Since VLMs can process and analyze data from multiple sources, they find use in areas like digital content creation, medical imaging, and self-driving technology.

The recent advancements in Multimodal Large Language Models (MLLMs) have further expanded the scope of AI in language and vision applications. These models can integrate textual and visual information seamlessly, providing capabilities like visual grounding, image generation and editing, and domain-specific multimodal analysis [3-10]. The ability of VLMs and MLLMs to bridge the gap between language and vision represents a significant milestone in the pursuit of artificial general intelligence.

Large VLMs are capable of generalizing efficiently across a diverse array of datasets and applications, including complex tasks that show impressive performance. They can understand documents, interpret images with instructions, and discuss visual content. What's more, VLMs have the ability to capture spatial characteristics in images; when identifying or segmenting subjects, locating items, or answering questions about their positions, they can generate bounding boxes or segmentation masks [11-18].

Our research emphasizes the significance of object detection in complex environments, using the advanced vision-language model, the Florence 2 model [18, 19]. In unstructured environments, regular object detection and classification can be difficult due to their complexity and variability. The Florence 2 model is particularly suitable for this type of task, thanks to its strong ability to understand both visual and textual information, and its fine-tuning capabilities [19]. To enhance the model's accuracy and efficiency, we train the Florence 2 model with data specific to the domain and adjust its parameters. As a result of these modifications, the model has been effectively optimized for object detection and is now performing comparably to detection models like YOLO [20].

The increasing prominence of advanced VLMs and MLLMs is also impacting the realm of academic writing. The power of these models lies in their ability to integrate and analyze information from various modalities, potentially revolutionizing how researchers approach literature reviews, data synthesis, and even the writing process itself [21-23] .

This paper demonstrates that the Florence2 model has a remarkable ability to adapt to a wide range of complex and unstructured settings. This adaptability makes it more reliable in situations where other models struggle. The fine-tuned version of Florence2 is not only more capable but also provides major advantages in complicated environments. Refs. [24] shows that VLMs can identify visual concepts from text-based prompts; however, Florence2 model proposed in this paper excels by integrating these capabilities effectively within complex and unstructured scenarios [24]. Through vision-language integration, the Florence2 model has become substantially enhanced. It improves object detection accuracy by providing a deeper and more contextual understanding of scenes. The model is adept at generalizing across various activities and situations, which ensures consistent performance even in unusual or dynamically changing environments. Refs. [25] point out the limitations of VLMs regarding granularity and specificity during zero-shot recognition. Florence2 addresses these concerns by being fine-tuned for specific domain tasks. Our refined Florence2 model offers numerous advantages over several state-of-the-art object detection models. Notably, it displays greater adaptability due to its multimodal capabilities, performs better in detailed environments, and is equipped to undertake additional tasks beyond just object detection. What's more, the VLM-PL model in [26] introduces a pseudo-labeling technique that enhances class-incremental object detection [26]. Given these advantages, Florence2 stands out as a strong and adaptable tool in computer vision, with the potential to greatly impact a variety of practical applications.

In this paper, we used the Florence 2 model after it underwent fine-tuning. We applied this fine-tuned model to object detection tasks in complex and unstructured environments, displaying its robustness and adaptability. These environments are characterized by a lack of structured data and clear boundaries, which present unique challenges. The Florence 2 model successfully navigated these difficulties. Our analysis indicates that the Florence 2 model performed almost on par with the YOLO, even surpassing the latest YOLOv10 model [20].

## 2. Recent Works

This section explores the key developments, current trends, and open challenges in object detection, highlighting the evolution from traditional methods to deep learning-based techniques and the future directions for this rapidly advancing field.

Object detection is a fundamental task in computer vision that focuses on recognizing and locating objects within an image or video frame. It is essential in applications such as autonomous driving [27], surveillance [28], robotics [29], healthcare [30] , and augmented reality [31]. Unlike image classification, which only assigns a label to an entire image, object detection identifies objects and provides their exact positions using bounding boxes or segmentation masks.

The development of object detection methods has progressed through several notable stages. Initial techniques were grounded in traditional machine learning and manually designed features. Early successes came from algorithms like the Viola-Jones Detector [32] and approaches that combined Histogram of Oriented Gradients (HOG) with Support Vector Machines (SVMs) [33]. These methods effectively captured fundamental visual patterns and paved the way for real-time object detection.

The advent of deep learning brought a paradigm shift to object detection. Convolutional Neural Networks (CNNs) enabled automated feature extraction, leading to significant improvements in detection accuracy and robustness. Models such as the R-CNN series [34], YOLO (You Only Look Once) [28], and SSD (Single Shot MultiBox Detector) [35] emerged as state-of-the-art solutions, each offering trade-offs between accuracy and speed. These advances have made real-time object detection feasible even in complex and dynamic environments [36].

Despite these achievements, several challenges persist, including the need to enhance detection accuracy, reduce computational costs, and ensure reliability across diverse real-world scenarios. Issues such as occlusion, varying lighting conditions, scale variation, and the need for real-time processing continue to drive ongoing research in the field [37, 38].

Recently, vision-language models have gained significant popularity and attention due to their ability to integrate visual and textual information for comprehensive understanding and prediction. Early vision-language models such as ViLBERT [39] and LXMERT [40] laid the groundwork for the field by introducing architectures that fused visual and linguistic representations effectively. ViLBERT proposed a two-stream transformer architecture where visual and textual information was processed separately before being fused [39]. This model showed substantial improvements in tasks like Visual Question Answering (VQA) and referring expressions. Similarly, LXMERT employed independent encoders for images and language, combining these representations through a cross-modality encoder, achieving strong performance in VQA and image retrieval tasks.

These models, while groundbreaking, were limited by their reliance on task-specific architectures and relatively small datasets. To overcome these limitations, [18] introduced the Florence model, a unified foundation model for vision-language understanding. Florence employs a single transformer-based architecture capable of handling multiple vision tasks, such as image classification, object detection, and segmentation. Its training on a large-scale dataset of 900 million images allows it to generalize well across different tasks and benchmarks like COCO [41] and LVIS [42]. Florence's unified approach simplifies the architecture by eliminating the need for task-specific models, enabling more efficient and scalable vision-language processing.

Building on the advancements of Florence, [43] introduced Florence-2, an improved vision-language model designed to push the boundaries further. Florence-2 incorporates enhanced pre-training techniques, such as contrastive learning and masked image modeling, and utilizes larger, more diverse datasets to achieve state-of-the-art performance. Florence-2 excels in zero-shot object detection, few-shot learning, and open-set recognition tasks, outperforming contemporary models like CLIP [44] and ALIGN [45]. These improvements make Florence-2 highly effective in real-world applications, including autonomous navigation, healthcare diagnostics, and robotics, where robustness to domain shifts and fine-grained visual understanding are essential.

## 3. Methodology

This paper investigates changes to the design of Florence 2 to improve its object recognition skills. The initial step involves analyzing the architecture of the Florence 2 model and its pre-training, with a focus on determining its strengths and weaknesses in object recognition. Next, during the data preparation phase, images, along with bounding boxes and labels, are loaded to create a variety of datasets. At this stage, the Florence 2 model is fine-tuned based on the identified features to enhance its effectiveness in driving real estate reform and innovation. Also, temporal fine-tuning is applied, using techniques like adaptive learning to adjust the weights of the previously trained models. Finally, the model's performance is measured using standard metrics such as Precision, Recall, and IoU to evaluate its accuracy and robustness.

### 3.1. Model Architecture

The Florence-2 model has a complex configuration, which can solve visual, and speech related problems. It uses Dual Attention Vision Transformer's (DaViT's) visual encoder to convert images into visual graphics before processing the text [46]. DaViT's visual encoder generates the model at four stages including a patch-embedding layer at the beginning of each stage. In the Florence-2 model, the visual embedding generated by DaViT are not directly added to BERT's text embedding; instead, visual and textual information are processed separately, and then transformer-based multi-modal encoders and decoders are used to produce the final output. The Florence2 tokenizer words include location tokens for function-specific fields. These tokens can refer to box shapes, such as top-left-right corners (x0, y0, x1, y1), or polygon vertices (x0, y0,…, xn, yn), and give the model the ability to a capturing and analyzing spatial data has improved .

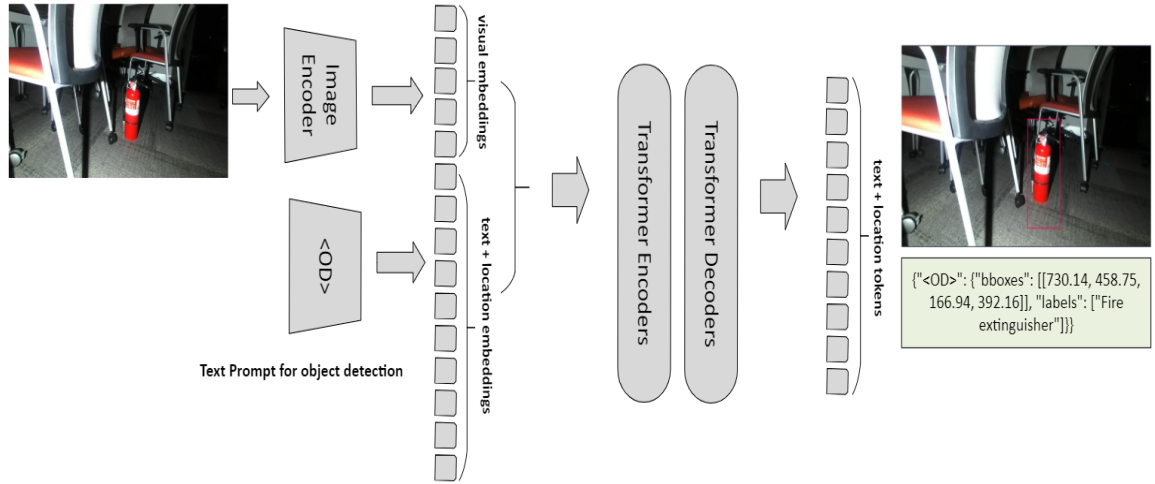

**Figure 1.** Architecture of the Florence2 model for object detection.

There are two versions of the Florence-2 series: Florence-2-base and Florence-2-large, with 0.23 billion and 0.77 billion parameters, respectively. Figure 1 shows the object detection process using the Florence2 model. The Image Encoder first processes the incoming image to perform the visual embedding. Search features are also built-in for identification, providing information and location. These embedding are then passed to Transformer Encoders and Decoders, which generate information and location tokens. The final results include bounding boxes and labels identifying objects in the image, as shown by the familiar "Fire extinguisher" on the right.

### 3.2. Data Description

The study uses the conjunction of the PST900 data set [47] and the DARPA SubT Autonomous Solution facility dataset [48]. Figures 2 and 3 show sample images from PST900 data set and the DARPA SubT Autonomous Solution facility, respectively. The focus is on the PST900-RGB subset, specifically designed for challenging conditions such as subterranean tunnels, mines, and caverns with restricted vision and illumination. Here's a thorough breakdown of the dataset, the dataset was created by collecting a wide range of photos from multiple sources apart from the images in the PST900 dataset, that were specifically designed for image classification, object identification, and image captioning applications. The final dataset comprises a total of 1788 images, with an equal contribution of 894 images from the PST900 dataset and 894 images from the DARPA SubT Autonomous Solution workspace. Figure 4 visualizes instances per class in the dataset. Table 1 summarizes object numbers in each object class. The combined dataset contains five object classes: Backpack (528 instances), Cellphone (259 instances), Drill (409 instances), Fire extinguisher (404 instances), and Survivor (537 instances).

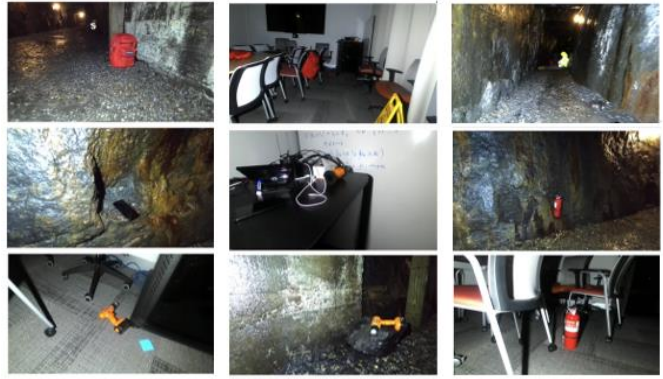

**Figure 2.** Sample images from PST900 data set [47].

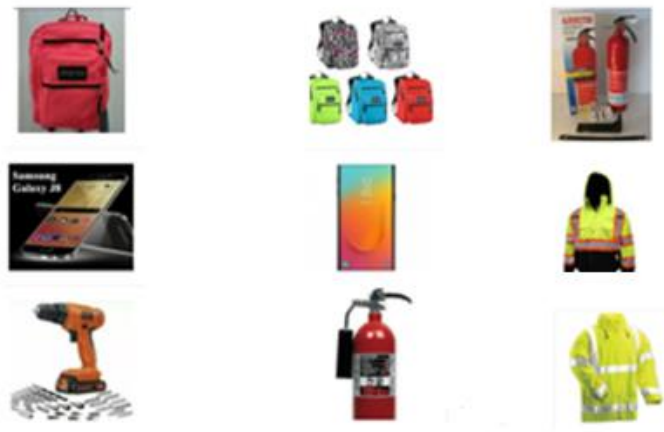

**Figure 3.** Sample images from the DARPA SubT Autonomous Solution facility data set [48].

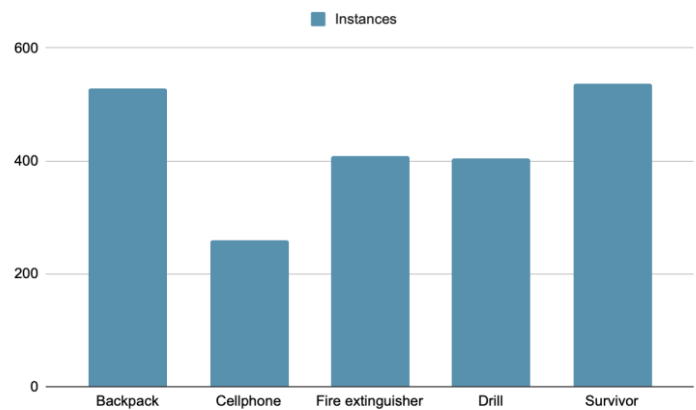

**Figure 4.** Instances per class in the dataset.

**Table 1.** Class distribution of dataset

| Class | Instances |
|:---:|:---:|
| Backpack | 528 |
| Cellphone | 259 |
| Drill | 409 |

| Fire extinguisher | 404 |
|---|---|
| Survivor | 537 |

**Table 2.** Dataset train/test/validation split.

| Set | Percentage | Number of images |
|---|---|---|
| Train set | 70% | 1251 |
| Validation set | 20% | 359 |
| Test set | 10% | 178 |

In the experiments, we flowed as shown in Data Processing Workflow in Figure 5. We first used RoboFlow platform for annotation and published our fused, improved, and augmented object detection data set in [49]. Table 2 presents the data number in training, testing, and validation settings. Later, several augmentation techniques in Table 3 were applied to take the image count to 2818 and 3981 respectively. The images were finally then split as in the workflow.

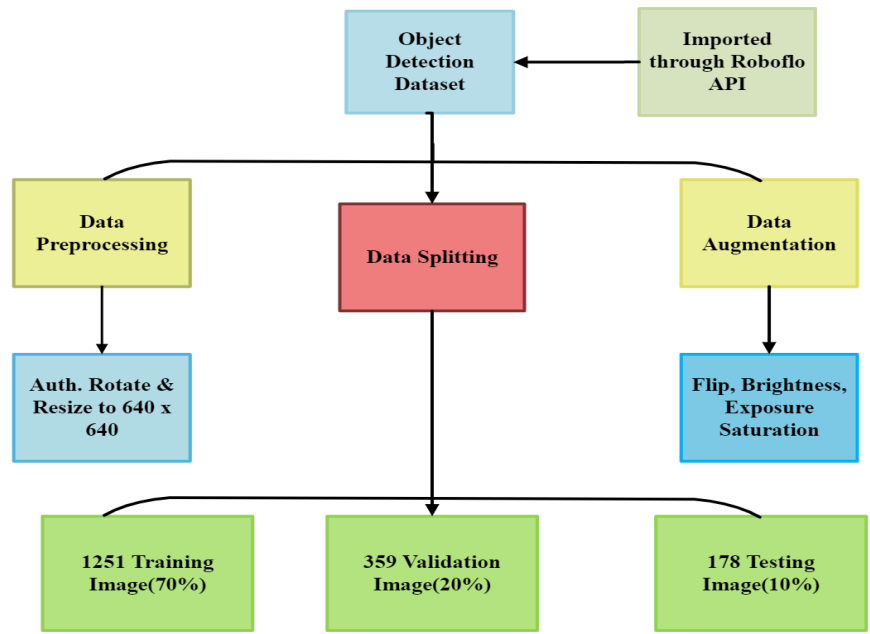

**Figure 5.** Data processing workflow for object detection.

**Table 3.** Data augmentation techniques.

| Technique | Details |
|---|---|
| Resize | Stretch to 640x640 |
| Flip | Horizontal |
| Rotate | 90° Clockwise, Counter-Clockwise |
| Shear | ±10° Horizontal, ±10° Vertical |
| Saturation | Between -25% and +25% |
| Exposure | Between -10% and +10% |

By implementing these enhancements, we boosted the diversity and robustness of the training dataset, allowing the Florence-2 model to generalize more effectively and perform better in real-world applications. Few examples of the unannotated and annotated images from the final dataset have been shown in the Figures 6 and 7. The first row shows a backpack, followed by cellphone, drill, fire extinguisher and finally survivor class

Following annotation and augmentation, the dataset was structured for Florence-2 using RoboFlow. This translation provides compatibility with Florence2 vision-language model, resulting in improved data interpretation and usage. Using Florence2 specialized format rather than a generic one like COCO takes advantage of the model's unique characteristics, increasing object detection accuracy and efficiency. This organized preparation provides Florence-2 with high-quality, well-annotated data that will help it perform better in tough circumstances.

Figure 6 shows sample annotated images from the dataset, displaying the labelled instances of classes such as backpack, drill, cellphone, survivor, and fire extinguisher.

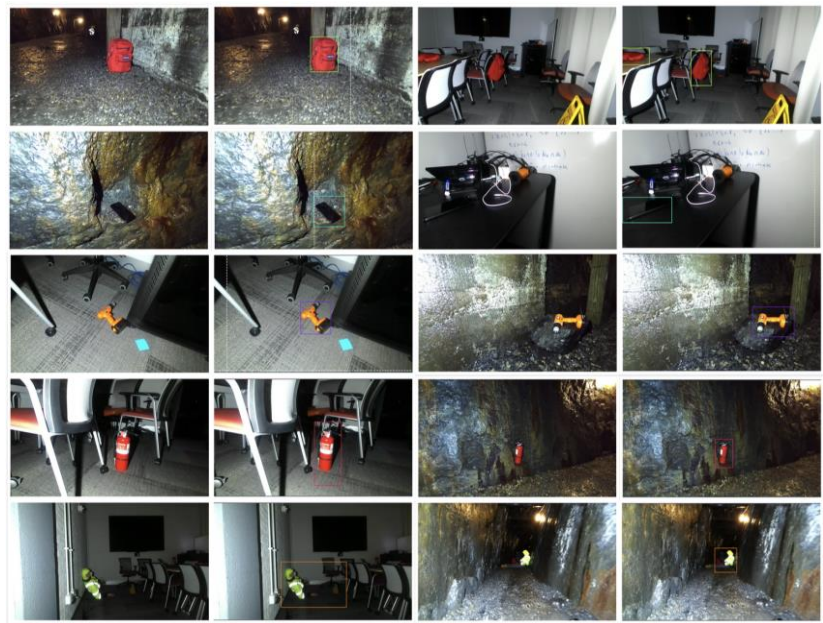

**Figure 6.** Sample images from the dataset demonstrating the variety of items and circumstances utilized in training, validation, and testing

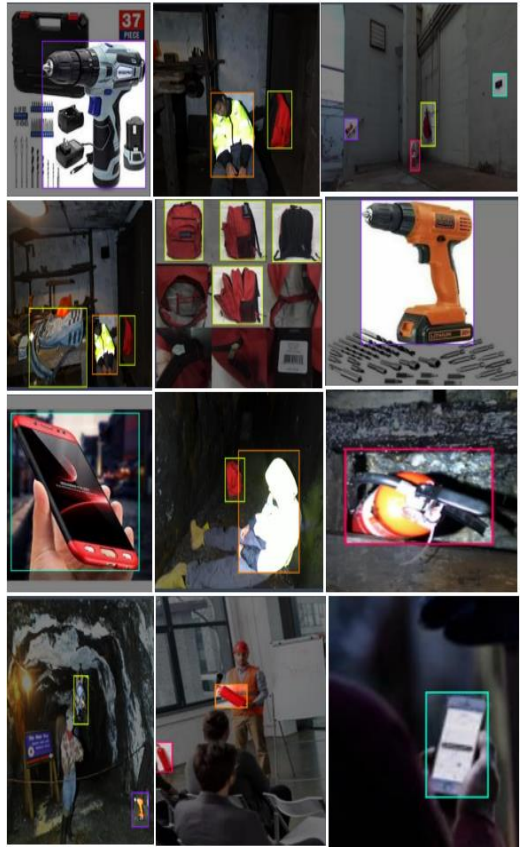

**Figure 7.** Sample annotated images from the dataset, displaying the labeled instances of classes such as backpack, drill, cellphone, survivor, and fire extinguisher.

### 3.3 Loading and Testing Pre-trained Florence-2 Model

Before fine-tuning, we loaded the pre-trained Florence-2 model into memory first, then fine-tuned it on the custom dataset. Florence-2 is available in two versions: base and big, with 230 million and 770 million parameters, respectively. In this paper, we used the base version to balance performance and resource requirements. After loading the model, we tested its inference abilities on a sample image. This step, while optional, acted as a check to confirm that our environment was properly configured.

### 3.4 Using LoRA to Optimize Florence 2 Training

To efficiently fine-tune the Florence-2 basic model, which contains 270 million parameters, we used Low-Rank Adaptation (LoRA) [50]. LoRA is especially beneficial for dealing with large models in resource-constrained contexts which has restricted computing resources and memory. LoRA improves the efficiency of the training process by reducing the number of parameters to be trained. Instead of updating the entire model, LoRA focuses on replacing only a fraction of its weight. This is done by adding a low-level decomposition step to the weight matrices of the model. LoRA modifies the model architecture by adding low-level trainable matrices to the original weight matrices, capturing important optimization information with fewer parameters. Full micro integration requires updating all model parameters, resulting in a larger (32-bit) optimizer environment and higher computational overhead. However, LoRA comes with low-cost (16-bit) adapt-ers embedded in the model, significantly reducing trainable parameters and computational resources while maintaining performance. In computation, LoRA splits the weight matrices (W) into two parts: the lower ranked matrix (L) and the adjustment matrix (A):

$$W' = W + \Delta W W' = W + \Delta W \tag{1}$$

$$W = LxA \tag{2}$$

where ΔW indicates that the original weight matrix has been minimally modified for W. The minimum matrices L and A are sufficiently smaller than the original weight matrix, reducing the computational cost and memory used for training up. This allows the model to detect significant changes without changing all the initial parameters. Using LoRA, we are able to update the Florence-2 model in a way that does not require large resources. This strategy not only simplifies the training process, but also improves the standard of change in certain activities by focusing on the main features (which are possible on limited hardware).

### 3.5. Fine-tuning Florence-2: Training Setup

We optimized the Florence-2 model at three-stages in training algorithm such as initialization, training machine, and validation loop. We tried several different settings to get the best results. Before starting the training process, we configured the optimizer and the number of classes. To help stabilize the training set, we used the AdamW optimizer, which is a modified version of Adam with constant weight loss and linear scheduler. The learning rate scheduler was set up to alter the learning rate dynamically during training. In addition, experimentations were also conducted with various LoRA parameters, such as ranks of 8, 4, 10, 16, and 32, alpha values of 8, 16, 32, and 20, dropout rates of 0.05 and 0.1. We spitted the low-rank matrices into multiple smaller subspaces (or ranks) during training and used Gaussian-initialized weights. During the training phase, we iterated through the dataset in batches, made forward passes, and calculated the loss. Backpropagation was used to change the model weights according to the calculated loss. The learning rate scheduler was adjusted to ensure accurate rate modifications. Following each training session, the model was tested on the validation set. This evaluation involved determining the loss without completing backpropagation, allowing us to gauge the model's performance and ensuring it generalizes well to previously unseen data. Table 4 shows our experimental settings for hyper parameter optimization

**Table 4.** Experimental settings for hyper parameter optimization

| Parameter | Values |
| --- | --- |
| Batch size | 3, 4, 6, 12 |
| Epochs | 7, 8, 10, 12 |
| Learning rate | 5e-6, 3e-6, 1e-6 |

| Optimizers | AdamW, AdamW (weight decay 0.01),m SGD, SGD (momentum 0.9, weight decay 0.01) |
| --- | --- |

The maximum training time for the fine-tuning process was approximately 4 hours on L4 GPU. Various batch sizes were tested out of which a batch size of 6 seemingly produced the optimum results. Epochs were also varied based on the available GPUs out of which epochs of around 10 on the L4 GPU gave the best results. More epochs resulted in overfitting. Learning rates ranging from 1e-6 to 5e-6 worked just fine. Using LoRA, we effectively reduced the number of trainable parameters to 1,929,928 out of 272,733,896, resulting in a 0.7076% trainable parameter rate. This reduction significantly optimized training efficiency and decreased the computational resources required.

### 3.6. Performance Analysis of Fine-tuned Florence-2

The mean average accuracy (mAP) metric was used to evaluate the performance of the optimized Florence-2 model, and the confusion matrix was evaluated. The mAP provides a detailed assessment of the accuracy of the model in several studies, among others very important for object recognition tasks. This is an important metric for evaluating object recognition algorithms because it represents the accuracy-recall trade-off between confidence levels and Intersection over Union (IoU) thresholds. IoU is a measure of overlap between the predicted bounding box and the ground truth bounding box. It is calculated as:

$$IoU = \frac{Area\ of\ Overlap}{Area\ of\ Union} \tag{3}$$

Unlike accuracy, which may not be sufficient for object recognition tasks, accuracy in multiple areas of mAP, IoU thresholds (estimate correct detection of objects within the predicted range) and recall (correctly identified within the predicted range). This enables a better assessment of the model performance, and it demonstrates its ability to accurately identify and position objects in images.

$$Precision = \frac{TP}{TP+FP} \tag{4}$$

where True Positive (TP) is the instances where the model correctly predicts the positive class. False Positive (FP) are the instances where the model incorrectly predicts the positive class.

$$Recall = \frac{TP}{TP+FN} \tag{5}$$

where FN (False Negative) is a metric used in classification and object detection tasks to represent instances where the model fails to correctly identify a positive case. A model with high recall successfully identifies most positive cases, ensuring minimal false negatives, whereas a model with low recall misses many positive cases, leading to a higher number of false negatives.

Precision-Recall (PR) Curves are generated by varying the reliability of each class. The area under these curves, known as Average Precision (AP), is then calculated to evaluate the performance of the model.

The mAP is calculated by averaging the AP values of all classes, which provides an integrated analysis of the detection performance. The area under the PR curve is mAP, which measures overall model performance.

$$mAP = \frac{1}{N} \sum_{i=1}^{N} AP_i \qquad (6)$$

where N is the number of classes and APi is the Average Precision for class *i*. mAP50 evaluates the accuracy of predictions with IoU ≥ 0.5. mAP50:95: considers IoU values ranging from 0.5 to 0.95 in increments (e.g., 0.05) and calculates the average across these values.

## 4. Experimental Results

We executed the experimental studies on Google Colab utilizing T4 and L4 GPUs by following the model training and adaptation. We followed the model training and adaptation workflow in Figure 8. We employed the study in two augmented datasets, consisting of 2818 and 3981 images respectively. Primarily, the dataset with 2818 images was utilized, considering resource constraints. The dataset with 3981 images was then tried. LoRA rank was set to 8, with an alpha parameter of 16, although additional trials were conducted with ranks of 4 and 16 for comparative analysis. The AdamW optimizer was predominantly used, alongside exploratory experiments with the Stochastic Gradient Descent (SGD) optimizer, both with and without weight decay. A learning rate scheduler was employed, adjusting the learning rate within the range of 3e-6 to 5e-6, identified as optimal for the dataset size. A batch size of 6 was maintained, and the number of epochs varied between 10 and 12. Epoch counts exceeding this range resulted in overfitting, whereas lower counts led to under fitting.

Prior to fine-tuning, the Florence2 model was tested for object detection on the same augmented dataset. The initial tests showed that the model was unable to detect the objects correctly, and in instances where it did detect objects, it failed to recognize the correct classes. Some examples of the detection results prior to fine-tuning are displayed in Figure 9. Not only it detected unnecessary classes but also some annotations were wrong or did not match with our requirements. Optimal results were attained with the medium-sized augmented dataset comprising 2818 annotated images, utilizing the Florence2-base-ft model. The most effective configuration included a learning rate of 5e-6, LoRA rank of 8, LoRA alpha of 16, and the AdamW optimizer without weight decay. This setup achieved a mAP at IoU = 0.50 (mAP50) of 0.80, with corresponding training and validation losses of 1.16 and 1.12, respectively. The detailed results and observations are presented in Table 5.

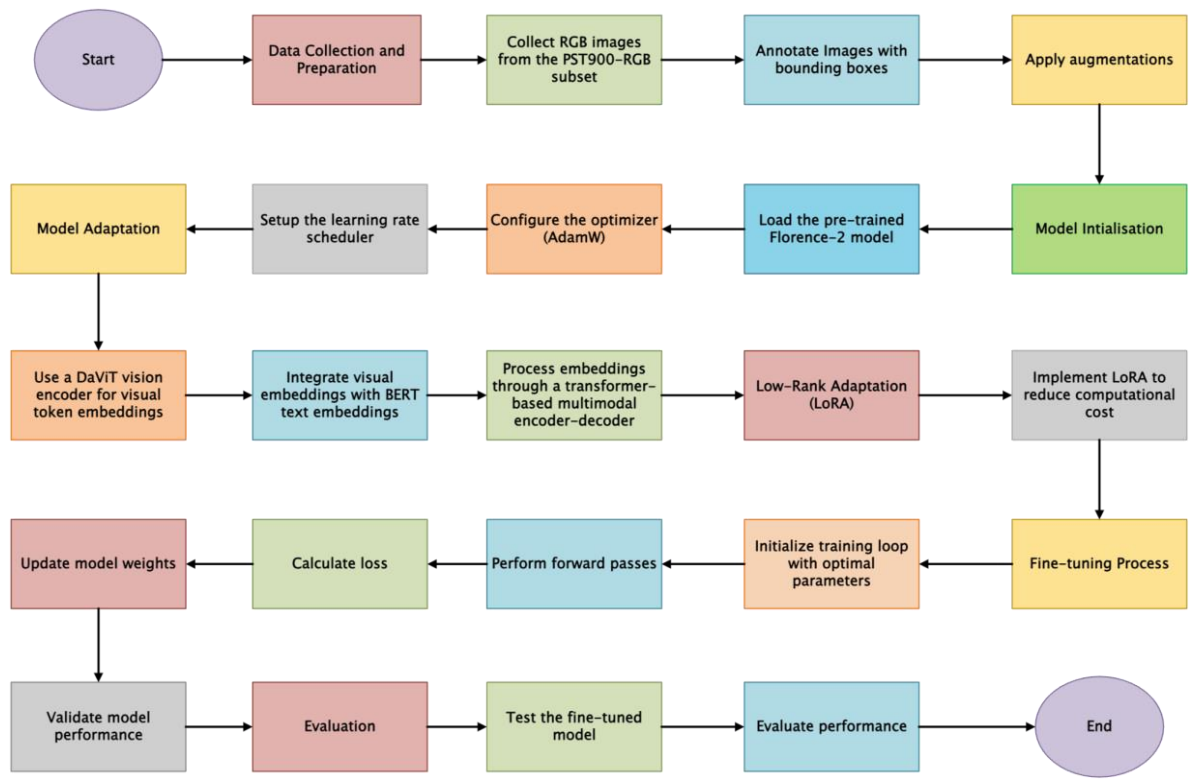

**Figure 8.** Overview of the model training and adaptation workflow

{"<OD>": {"bboxes": [[125.1199951171875, 382.3999938964844, 509.7599792480469], [486.7200012207031, 639.0399780273438, 639.0399780273438], [152.0, 383.03997802734375, 224.95999145507812], [578.239990234375, 639.0399780273438, 331.1999816894531], [125.1199951171875, 338.8800048828125, 510.3999938964844]], "labels": ["backpack", "chair", "chair", "chair", "suitcase"]}}

{"<OD>": {"bboxes": [[249.27999877929688, 18.8799991607666, 438.0799865722656, 577.5999755859375], [214.0800018310547, 227.51998901367188, 576.9599609375, 639.0399780273438]], "labels": ["mobile phone", "person"]}}

{"<OD>": {"bboxes": [...]}}

{"<OD>": {"bboxes": [[109.75999450683594, 43.84000015258789, 615.3599853515625, 589.1199951171875]], "labels": ["toy"]}}

**Figure 9.** Predictions of the model prior to fine-tuning.

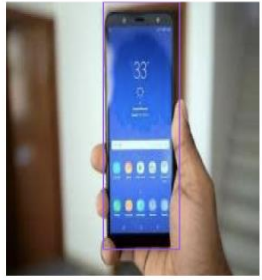

{"<OD>": {"bboxes": [[246.0800018310547, 0.3199999928474426, 438.0799865722656, 578.8800048828125]], "labels": ["Cellphone"]}}

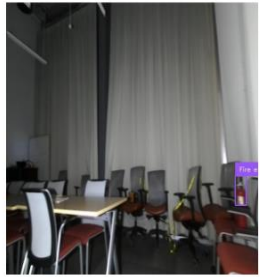

{"<OD>": {"bboxes": [[582.0800170898438, 407.3599853515625, 609.5999755859375, 476.47998046875]], "labels": ["Fire extinguisher"]}}

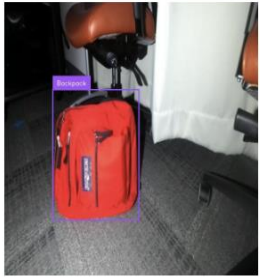

{"<OD>": {"bboxes": [[122.55999755859375, 205.1199951171875, 339.5199890136719, 510.3999938964844]], "labels": ["Backpack"]}}

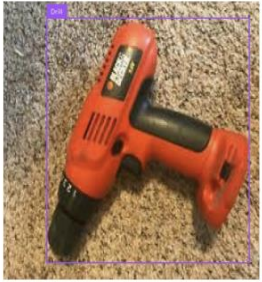

{"<OD>": {"bboxes": [[109.1199951171875, 38.07999801635742, 614.719970703125, 601.9199829101562]], "labels": ["Drill"]}}

**Figure 10.** Predictions of the model after fine-tuning.

As demonstrated in Table 5, the optimal performance was observed with a LoRA rank of 8, the AdamW optimizer, and a learning rate within the estimated optimal range, aligning well with established theories on fine-tuning LLMs. Figure 10 shows predictions of the model after fine-tuning.

For further evaluation of the fine-tuned Florence2 model, state-of-the-art YOLO models, specifically Yolov8, Yolov9, and Yolov10, were trained to perform object detection on the same augmented dataset of 2818 images. These models were configured with a batch size of 6 and trained for 12 epochs, mirroring the fine-tuning parameters used for the Florence2 model. The comparative results of these evaluations are presented in Table 6.

Table 5. Performance metrics of the model across different configurations.

| Sl. | GPU | Images | LR | Optimizer | LoRA | | Metrics | | | | |
|---|---|---|---|---|---|---|---|---|---|---|---|
| | | | | | Rank | Alpha | mAP50 | mAP50_95 | mAP75 | Training Loss | Validation loss |
| 1 | L4 | 2818 | 5e-06 | AdamW | 8 | 16 | 0.80 | 0.57 | 0.56 | 1.16 | 1.12 |
| 2 | L4 | 2818 | 5e-06 | SGD | 8 | 16 | 0.60 | 0.44 | 0.42 | 1.41 | 1.29 |
| 3 | L4 | 2818 | 1e-06 | AdamW (weight decay = 0.01) | 4 | 8 | 0.56 | 0.39 | 0.36 | 1.44 | 1.30 |
| 4 | A100 | 2818 | 5e-06 | AdamW (weight decay = 0.01) | 16 | 32 | 0.66 | 0.46 | 0.45 | 1.12 | 1.03 |
| 5 | T4 | 3981 | 3e-06 | SGD (momentum = 0.9) | 8 | 16 | 0.74 | 0.52 | 0.50 | 1.41 | 1.32 |
| 6 | L4 | 3981 | 3e-06 | AdamW | 8 | 16 | 0.79 | 0.57 | 0.54 | 1.17 | 1.08 |

Table 6. Performance comparison of YOLO models.

| Model | GPU | Images | mAP50 | mAP50_95 |
|---|---|---|---|---|
| Yolov8 | T4 | 2818 | 0.84 | 0.56 |
| Yolov9 | T4 | 2818 | 0.84 | 0.58 |
| Yolov10 | T4 | 2818 | 0.74 | 0.48 |

As evidenced by the results, the fine-tuned Florence2 model has notably outperformed the latest Yolov10 model in terms of object detection accuracy. However, it marginally lags behind the Yolov8 and Yolov9 models. Nevertheless, considering the extensive array of capabilities retained by the fine-tuned model—such as image captioning, detailed captioning, semantic segmentation, and caption-to-phrase grounding—these results are highly commendable. Given the ongoing research in fine-tuning methodologies and the emergence of innovative approaches, coupled with the immense potential inherent in vision-language models, the Florence2 model holds promise for achieving superior performance. It is anticipated that with further refinements, the Florence2 model will surpass all current YOLO models in object detection efficacy.

The training and validation loss curves are depicted in Figures 11 and 12, respectively. Figure 13 presents the confusion matrix, providing a comprehensive understanding of the fine-tuned model's performance in accurately detecting objects across various classes in the validation set.

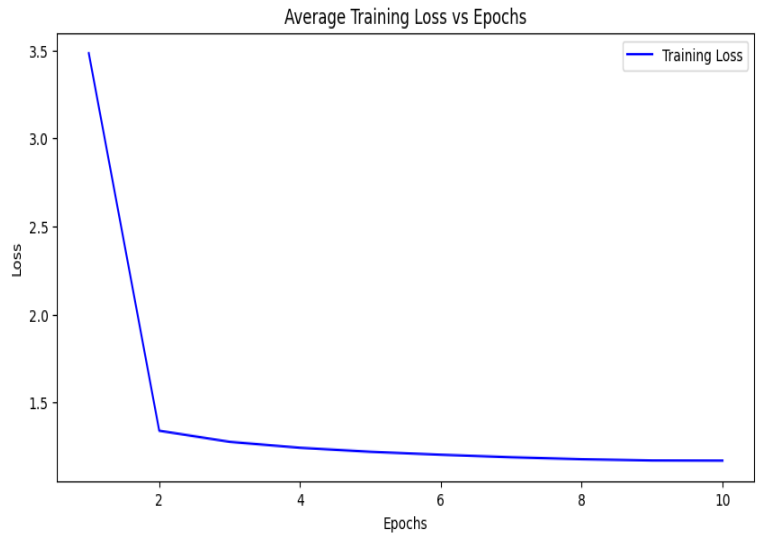

**Figure 11.** Average training loss across epochs during model training

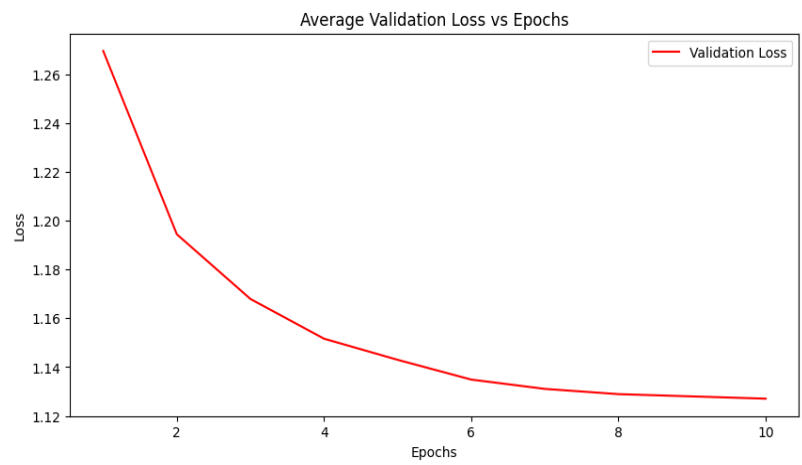

**Figure 12.** Validation loss trend over epochs during model training

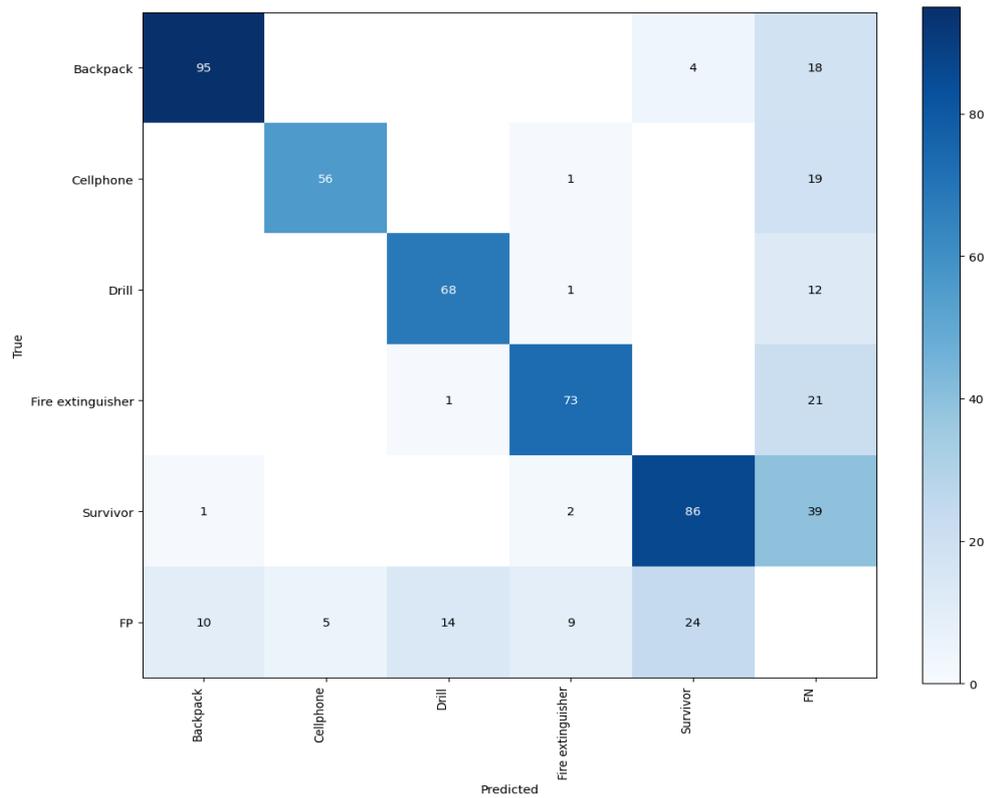

**Figure 13.** Confusion matrix illustrating the model's classification performance across different classes.

The confusion matrix in figure 16 shows that the fine-tuned Florence2 model has successfully classified the detected objects, with very few instances of misclassification. Continuous advancements in sophisticated techniques and fine-tuning processes have the potential to enhance the performance of the Florence2 model in computer vision tasks, particularly in object detection. Further, ongoing research and development in this field are likely to lead to improvements in the models' accuracy and consistency, thereby increasing their applicability across various contexts in computer vision.

## 5. Conclusions

The goal of this study is to illustrate the process of fine-tuning the Florence2 vision language model for performing computer vision tasks, particularly object detection environments that are denied of GPS, complex and unstructured. Our thorough experimentation and analyses yielded satisfactory results, showing that the fine-tuned Florence2 model performs comparably to the most advanced object detection models. The outcomes from our experiments emphasize the effectiveness of the Florence2 model in the domain of object detection. Also, this model is a superior alternative to certain models and algorithms specifically developed solely for object detection tasks, like YOLO. This is because of the model's additional capabilities, which extend beyond basic object detection to include functions such as visual question answering, detailed captioning, and caption-to-phrase grounding, among others. Such features emphasize the model's versatility, rendering it a valuable asset for a diverse array of computer vision tasks. Besides, this demonstrates how vision language models can be used for multitasking applications, even after being fine-tuned for excellence in a specific domain. Through a series of optimization and experimentation procedures, we identified the optimal configuration, which employs the AdamW optimizer, a learning rate between 3e-6 and 5e-6, and a LoRA rank of 8 with a LoRA alpha of 16. This configuration, coupled with a batch size of 6 and a duration of 10 to 12 epochs on the available L4 GPU, produced the most favorable results, as discussed in the results section. A comparative analysis with leading YOLO models revealed that the fine-tuned Florence2 model exceeded the accuracy of Yolov10 and achieved competitive results with both Yolov8 and Yolov9 models. According to the confusion matrix analysis and loss curves, the model correctly classified the majority of items. What's more, there is potential for further performance enhancement through the integration of higher-quality images and a larger dataset for the model to learn from. In summary, this research demonstrates the effective fine-tuning of the Florence2 model for object detection, achieving a level of performance that surpasses some existing computer vision models while retaining its additional functionalities. This study emphasizes the potential and feasibility of using vision-language models for a wide range of complex tasks, thereby contributing to the advancement of the computer vision field. Prospective advancements in this domain suggest the likelihood of even greater achievements, positioning vision-language models as essential tools for research and applications in computer vision.

State-of-the-art methods structured for computer vision tasks, like YOLO, have set high benchmarks for object detection. YOLO models are well-known for their precision and speed, which makes them ideal for real-time applications. The fine-tuned Florence2 model outperformed the latest YOLOv10 model, but still lags slightly behind the YOLOv9 and YOLOv8 models in a few metrics, which shows there is still room for improvement. This can potentially be achieved with higher quality datasets and trying out latest techniques of fine-tuning that are under research at the moment. The vision language model still retains strong additional features which can all be helpful in real-world applications beyond object recognition and can be useful for multitasking purposes. This study does have certain drawbacks, though. A notable limitation was the computational resources at hand, which impacted the selection of dataset dimensions and the degree of hyper parameter adjustment. The dataset versions used contained only 2818 and 3981 images, which can be enough for initial assessment but can miss out on several details which are critical especially in cases of GPS-denied environments. Also, the complexity of unstructured environments presents difficulties that require more experimentation and fine-tuning of the model. Research on vision language models is advancing quickly, but there is a notable lack of studies addressing their effectiveness in unstructured, GPS-denied settings. This study intends to bridge that gap by displaying the capabilities of Florence2 in such contexts. The comparison with YOLO models emphasizes the potential of vision language models to provide superior adaptability and performance in complex scenarios. Though the model was fine-tuned for object detection, in which its performance improved drastically, it still retained its additional features which offer a more comprehensive contextual knowledge, increasing the model's usefulness in several applications. Because of its adaptability, Florence2 may be used for a variety of practical tasks, such as GPS tracking and autonomous navigation for robots. Potential future work may involve experimenting with higher resolution datasets and more advanced augmentation techniques. Further exploration into adaptive learning rate schedules and advanced optimization algorithms can yield even better results. The integration of novel fine-tuning strategies, such as self-supervised learning and transfer learning from related tasks, may also enhance the model's performance.

**Funding:** This research was funded by The Scientific and Technological Research Council of Türkiye, TÜBİTAK, grant number 123E406 and was funded by FIRAT University Scientific Research Projects Unit, FUBAP, grant number MF.24.80.